\documentclass[runningheads]{llncs}
\usepackage{amsmath}
\usepackage{amssymb}
\usepackage{amsfonts}
\usepackage{graphicx}
\usepackage{float}
\usepackage{subfigure}
\usepackage{booktabs}
\usepackage{multirow}
\usepackage{diagbox}
\usepackage{stfloats}
\usepackage{cite}
\usepackage{url}
\usepackage{xspace}
\usepackage{siunitx}
\usepackage[misc]{ifsym}
\usepackage{hyperref}
\hypersetup{
    colorlinks=true,
    linkcolor=blue,
    filecolor=blue,
    urlcolor=blue,
    citecolor=blue,
}

\makeatletter

\newcommand{\Rmnum}[1]{\expandafter\@slowromancap\romannumeral #1@}
\makeatother

\begin{document}
\title{Self-Supervised Learning for MRI Reconstruction with a Parallel Network Training Framework}
\titlerunning{Self-Supervised MRI Reconstruction}
\author{Chen Hu\inst{1, 2} \and Cheng Li\inst{1} \and Haifeng Wang\inst{1} \and Qiegen Liu\inst{5} \and Hairong Zheng\inst{1} \and Shanshan Wang\inst{1, 3, 4 (\textrm{\Letter})}}
\authorrunning{C. Hu et al.}
\institute{Paul C. Lauterbur Research Center for Biomedical Imaging, Shenzhen Institutes of Advanced Technology, Chinese Academy of Sciences, Shenzhen, Guangdong, China
\email{sophiasswang@hotmail.com; ss.wang@siat.ac.cn}
\and University of Chinese Academy of Sciences, Beijing, China
\and Peng Cheng Laboratory, Shenzhen, Guangdong, China
\and Pazhou Lab, Guangzhou, Guangdong, China
\and Department of Electronic Information Engineering, Nanchang University, Nanchang, Jiangxi, China
}

\maketitle
\begin{abstract}
Image reconstruction from undersampled k-space data plays an important role in accelerating the acquisition of MR data, and a lot of deep learning-based methods have been exploited recently. Despite the achieved inspiring results, the optimization of these methods commonly relies on the fully-sampled reference data, which are time-consuming and difficult to collect. To address this issue, we propose a novel self-supervised learning method. Specifically, during model optimization, two subsets are constructed by randomly selecting part of k-space data from the undersampled data and then fed into two parallel reconstruction networks to perform information recovery. Two reconstruction losses are defined on all the scanned data points to enhance the network’s capability of recovering the frequency information. Meanwhile, to constrain the learned unscanned data points of the network, a difference loss is designed to enforce consistency between the two parallel networks. In this way, the reconstruction model can be properly trained with only the undersampled data. During the model evaluation, the undersampled data are treated as the inputs and either of the two trained networks is expected to reconstruct the high-quality results. The proposed method is flexible and can be employed in any existing deep learning-based method. The effectiveness of the method is evaluated on an open brain MRI dataset. Experimental results demonstrate that the proposed self-supervised method can achieve competitive reconstruction performance compared to the corresponding supervised learning method at high acceleration rates (4 and 8). The code is publicly available at \url{https://github.com/chenhu96/Self-Supervised-MRI-Reconstruction}.
\end{abstract}

\section{Introduction}
Magnetic resonance imaging (MRI) becomes an essential imaging modality in clinical practices thanks to its excellent soft-tissue contrast. Nevertheless, the inherently long scan time limits the wide employment of MRI in various situations. Acquiring MR data at sub-Nyquist rates followed by image reconstruction is one of the common approaches for MRI acceleration. However, image reconstruction is an ill-posed inverse problem, and high acceleration rates might lead to noise amplification and residual artifacts in the images \cite{joint-sparse, wangreview}. Therefore, recover high-quality MR images from undersampled data is a meaningful but challenging task.

In the past few years, promising performance has been achieved in deploying deep learning-based methods for MRI reconstruction \cite{shanshanwang16, KIKI-Net}. These methods can be broadly divided into two categories: data-driven networks and physics-based unrolled networks. The former can be described as training pure deep neural networks to learn the nonlinear mapping between undersampled data / corrupted images and fully-sampled data / uncorrupted images. Representative works include U-Net \cite{U-Net}, GANCS \cite{GANCS}, etc \cite{DeepcomplexMRI}. Unrolled networks construct network architectures by unfolding Compressive Sensing (CS) algorithms. Examples of this category are ISTA-Net \cite{ISTA-Net}, ADMM-Net \cite{ADMM-Net}, MoDL \cite{MoDL}, etc \cite{VN}. Regardless of the approaches utilized, most existing deep learning-based methods rely on fully-sampled data to supervise the optimization procedure. However, it is difficult to obtain fully-sampled data in many scenarios due to physiological constraints or physical constraints. Recently, a self-supervised learning method (self-supervised learning via data undersampling, SSDU) was proposed specifically to solve the issue \cite{self-supervised}, where the undersampled data is split into two disjoint sets. One is treated as the input and the other is used to define the loss. Despite the impressive reconstruction performance achieved, there are two important issues. First, the two sets need to be split with caution. When the second set does not contain enough data, the training process becomes unstable. Second, since no constraint is imposed on the unscanned data points, there is no guarantee that the final outputs are the expected high-quality images and high uncertainties exist.

To address the above issues, we propose a novel self-supervised learning method with a parallel network training framework. Here, differently, we construct two subsets by randomly selecting part of k-space data from the undersampled data, and then feed them into two parallel networks. Accordingly, two reconstruction losses are defined using all of the undersampled data to facilitate the networks' capability of recovering the frequency information and ensure that stable model optimization can be achieved. In addition, a difference loss is introduced, which acts as an indirect and reasonable constraint on the unscanned data points, between the outputs of the two networks to better aid in the subsequent high-quality image reconstruction during model testing. In the test phase, the obtained undersampled data is fed to either of the two trained networks to generate the high-quality results. Our major contributions can be summarized as follows: 1) A parallel network training framework is constructed to accomplish self-supervised image reconstruction model development through recovering undersampled data. 2) A novel difference loss on the unscanned data points of the undersampled data is introduced with the parallel networks, which can effectively constrain the solution space and improve the reconstruction performance. 3) Our method outperforms the existing state-of-the-art self-supervised learning methods and achieves a reconstruction performance competitive to the corresponding supervised learning method at high acceleration rates on an open IXI brain scan MRI dataset.
\begin{figure}[!htbp]
    \centering
    \subfigure[Training phase]{
    \includegraphics[width=\textwidth]{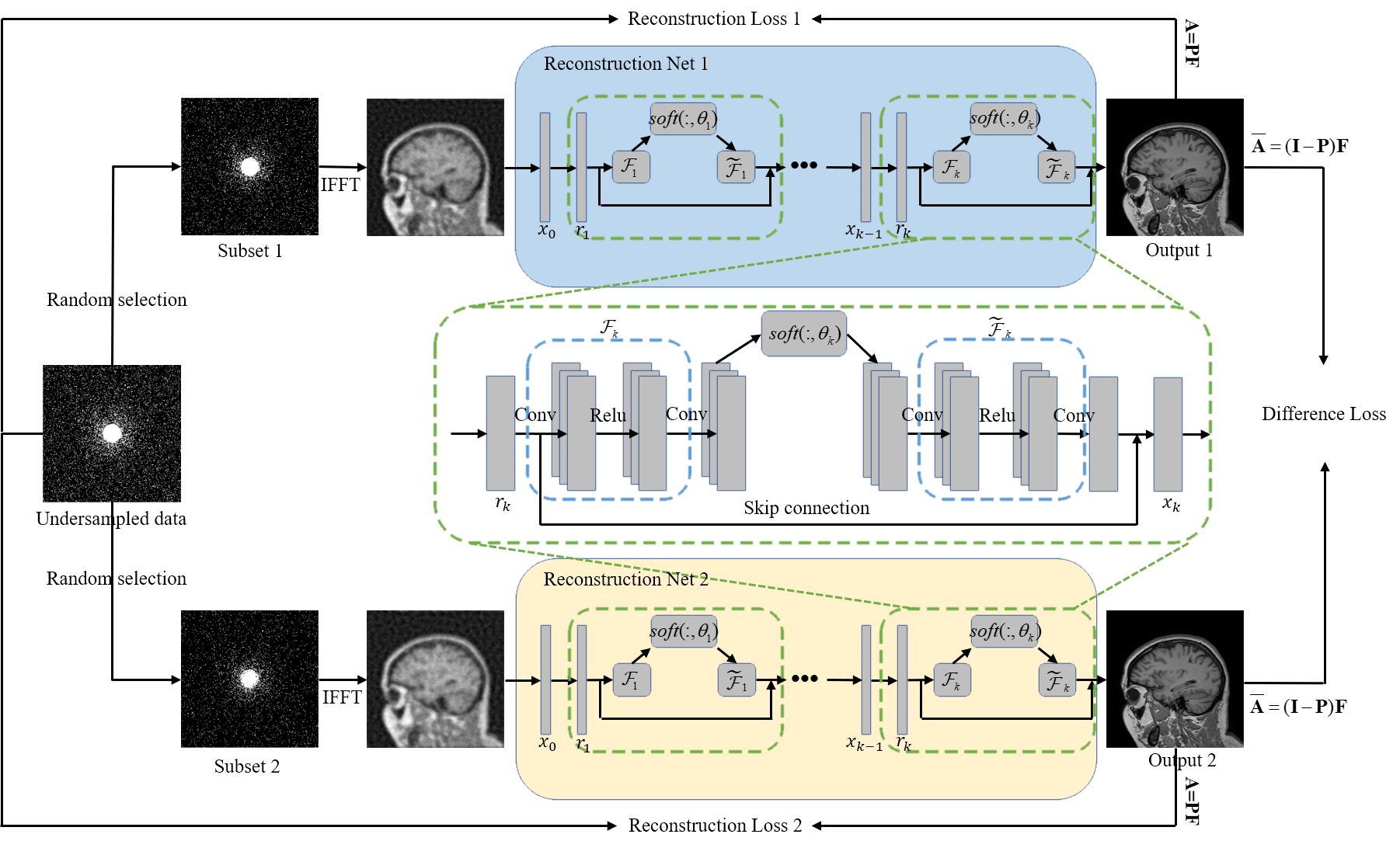}
    }
    \subfigure[Test phase]{
    \includegraphics[width=0.8\textwidth]{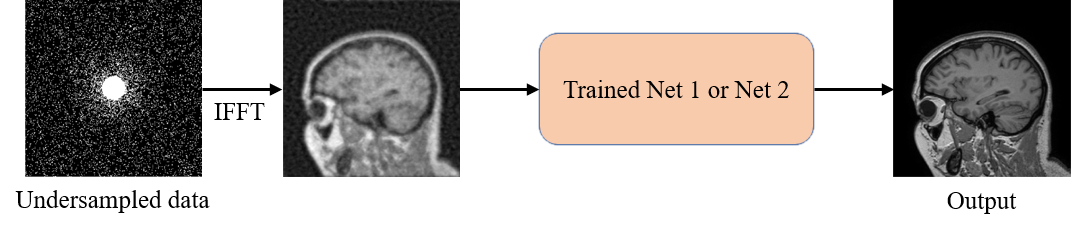}
    }
    \caption{The pipeline of our proposed framework for self-supervised MRI reconstruction.}
    \label{fig.1}
\end{figure}

\section{Methods}
Theoretically, the proposed method can be integrated with any existing deep learning-based method. In this work, an unrolled network, ISTA-Net \cite{ISTA-Net}, is utilized. The details are demonstrated as follows. 

\subsection{Mathematical Model of CS-MRI Reconstruction}
Mathematically, the CS-MRI reconstruction problem can be written as:
\begin{equation}\label{eq.1}
    \underset{\mathbf{x}}{\arg \min} \frac{1}{2} \| \mathbf{A} \mathbf{x} - \mathbf{y} \|_{2}^{2} + \lambda R(\mathbf{x})
\end{equation}
where $\mathbf{x}$ is the desired image, $\mathbf{y}$ is the undersampled k-space measurement, $\mathbf{A}$ denotes the encoding matrix which include Fourier transform $\mathbf{F}$ and sampling matrix $\mathbf{P}$, $R(\mathbf{x})$ denotes the utilized regularization, and $\lambda$ is the regularization parameter. The purpose of MRI reconstruction is to recover the desired image $\mathbf{x}$ from its measurement $\mathbf{y}$.

\subsection{Brief Recap of ISTA-Net}
ISTA-Net is an unrolled version of the Iterative Shrinkage Thresholding Algorithm (ISTA) \cite{ISTA-Net}, for which the regularization term in Eq. (\ref{eq.1}) is specified to be the L1 regularization. ISTA-Net solves the inverse image reconstruction problem in Eq. (\ref{eq.1}) by iterating the following two steps:
\begin{eqnarray}
    \mathbf{r}^{(k)} & = & \mathbf{x}^{(k-1)} - \rho \mathbf{A}^{\top} \left( \mathbf{A} \mathbf{x}^{(k-1)} - \mathbf{y} \right) \label{eq.2}\\
    \mathbf{x}^{(k)} & = & \widetilde{\mathcal{F}}(soft(\mathcal{F}(\mathbf{r}^{(k)}), \theta)) \label{eq.3}
\end{eqnarray}
where $k$ is the iteration index, $\rho$ is the step size, $\mathcal{F}(\cdot)$ denotes a general form  of image transform, $\widetilde{\mathcal{F}}(\cdot)$ denotes the corresponding left inverse, $soft(\cdot)$ is the soft thresholding operation, and $\theta$ is the shrinkage threshold. $\mathcal{F}(\cdot)$ and  $\widetilde{\mathcal{F}}(\cdot)$ are realized through neural networks and $\widetilde{\mathcal{F}}(\cdot)$ has a structure symmetric to that of $\mathcal{F}(\cdot)$. All free parameters and functions can be learned by end-to-end network training. More details can be find in \cite{ISTA-Net}.

\subsection{Proposed Self-Supervised Learning Method}
Fig. \ref{fig.1} shows the overall pipeline of our proposed framework for self-supervised MRI reconstruction. It includes a training phase of network optimization with only undersampled data and a test phase of high-quality image reconstruction from undersampled data.

\begin{figure}[!htbp]
    \centering
    \includegraphics[width=\textwidth]{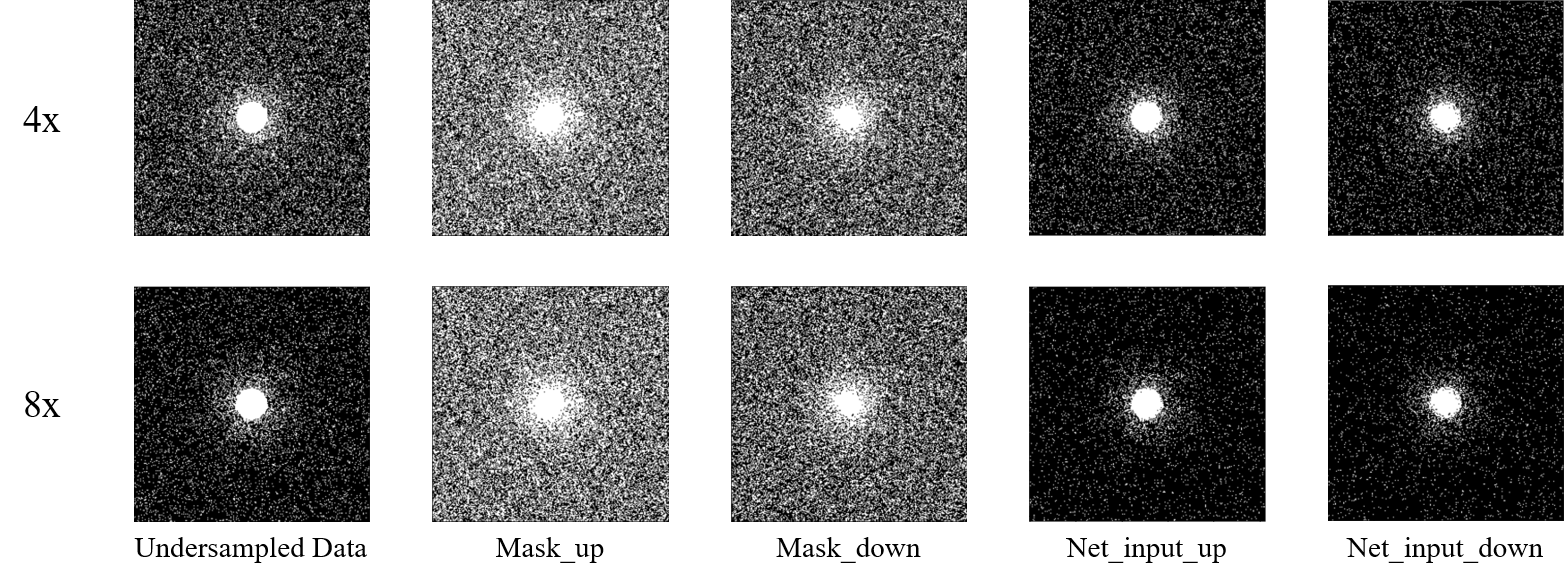}
    \caption{Construction of the two subsets. The autocalibrating signal (ACS) lines of the undersampling mask and the selection mask are 24 and 16, respectively. 2D random undersampling is utilized for both the undersampled data generation and the subsets construction. ``up'' and ``down'' refer to the two parallel networks.}
    \label{fig.2}
\end{figure}

In the training phase, two subsets are constructed by randomly selecting part of k-space data from the undersampled data, and then fed into two parallel networks. In this work, the reconstruction network utilizes ISTA-Net$^+$, which is an enhanced version of ISTA-Net \cite{ISTA-Net}. The architecture of ISTA-Net$^+$ is showed in Fig. \ref{fig.1}, and the number of iterations is set to 9. As illustrated in Fig. \ref{fig.2}, we construct the two subsets by taking the intersections of the undersampling mask and selection masks. The following strategies are adopted when choosing the selection masks: 1) The selection masks used by the two parallel networks should be different. 2) The input to the networks should include most of the low-frequency data points and part of the high-frequency data points. 3) We keep the numbers of the selected data points to be roughly half of the number of the undersampled data points. Two masks which are similar to the undersampling pattern are chosen as our selection masks. During network optimization, the undersampled data are used to calculate the reconstruction loss. Furthermore, a difference loss is defined to impose an indirect constraint to the unscanned data points, which can ensure that the learned unscanned data points of the two parallel networks are consistent. It is expected that the reconstructed images of the two networks are roughly the same since they are basically recovering the same thing. Overall, the network training process solves the following optimization problem:
\begin{equation}
    \underset{\mathbf{x}_1, \mathbf{x}_2}{\arg \min} \frac{1}{2} \| \mathbf{A}_1 \mathbf{x}_1 - \mathbf{y}_1 \|_2^2 + \frac{1}{2} \| \mathbf{A}_2 \mathbf{x}_2 - \mathbf{y}_2 \|_2^2 + \lambda R(\mathbf{x}_1) + \mu R(\mathbf{x}_2) + \nu \mathcal{L}(\mathbf{x}_1 , \mathbf{x}_2)
\end{equation}
where $\mathbf{y}_1$ and $\mathbf{y}_2$ are the two subsets selected from the undersampled k-space data, $\mathbf{A}_1$ and $\mathbf{A}_2$ are the corresponding encoding matrices, and $\mathbf{x}_1$ and $\mathbf{x}_2$ are the two estimations of the ground-truth image $\mathbf{x}$ which should be theoretically consistent. $R(\cdot)$ is the regularization. $\mathcal{L}(\cdot)$ denotes some similarity metrics. $\lambda$, $\mu$, and $\nu$ are the regularization parameters.

In the test phase, high-quality MR images are generated by inputting the raw undersampled data into either of the two trained networks. The network is expected to be generalizable enough and to be able to self-speculate the missing k-space data points.

\subsection{Implementation Details}
The loss function for model training is:
\begin{equation}
    \begin{aligned}
        \mathcal{L}(\Theta) & = \frac{1}{N} \left( \sum_{i=1}^N \mathcal{L}(\mathbf{y}^i, \mathbf{A}^i \mathbf{x}_1^i) + \sum_{i=1}^N \mathcal{L}(\mathbf{y}^i, \mathbf{A}^i \mathbf{x}_2^i)
        \right. \\
        & \left. + \alpha \sum_{i=1}^N \mathcal{L}(\overline{\mathbf{A}^i} \mathbf{x}_1^i, \overline{\mathbf{A}^i} \mathbf{x}_2^i) + \beta \mathcal{L}_\text{cons1} + \gamma \mathcal{L}_\text{cons2} \right)
    \end{aligned}
\end{equation}
where $N$ is the number of the training cases, $i$ denotes the $i^{th}$ training case. $\mathbf{x}_k = f(\mathbf{y}_k, \mathbf{A}_k; \theta_k), k = 1, 2$, where $f(\cdot)$ denotes the reconstruction network specified by the parameter set $\theta$. $\mathbf{A} = \mathbf{P} \mathbf{F}$ and $\overline{\mathbf{A}} = (\mathbf{I} - \mathbf{P}) \mathbf{F}$ refer to the scanned k-space data points and the unknown / unscanned k-space data points which are utilized to calculate the reconstruction loss and the difference loss, respectively. $\mathcal{L}_\text{cons}$ denotes the constraint loss in \cite{ISTA-Net}, which is included to ensure the learned transform $\mathcal{F}(\cdot)$ satisfies the symmetry constraint $\widetilde{\mathcal{F}} \circ \mathcal{F} = \mathcal{I}$. $\alpha$, $\beta$ and $\gamma$ are the regularization parameters. In our experiments, the loss metric $\mathcal{L}(\cdot)$ is set to the mean square error (MSE) loss, and $\alpha = \beta = \gamma = 0.01$.

The proposed method is implemented in PyTorch. We use Xavier \cite{Xavier} to initialize the network parameters with a gain of 1.0. To train the parallel networks, we use Adam optimization \cite{Adam} with a learning rate warm up strategy (the learning rate is set to 0.0001 after 10 warm up epochs) and a batch size of 4. The learning rate is automatically reduced by a constant factor when the performance metric plateaus on the validation set. All experiments are conducted on an Ubuntu 18.04 LTS (64-bit) operating system utilizing two NVIDIA RTX 2080 Ti GPUs (each with a memory of 11 GB).

\section{Experiments and Results}
\subsubsection{Dataset.}
The open-source dataset, Information eXtraction from Images (IXI), was collected from three hospitals in London. For each subject, T1, T2, PD-weighted, MRA, and diffusion-weighted images are provided. More details including scan parameters can be found on the official website\footnote{http://brain-development.org/ixi-dataset/}. Our experiments are conducted with 2D slices extracted from the brain T1 MR images, and the matrix size of each image is $256 \times 256$. The training set, validation set, and test set for our experiments contain 850 slices, 250 slices, and 250 slices, respectively. The image intensities are normalized to $[0, 1]$ before the retrospective undersampling process.
\begin{table}[!htbp]
    \centering
    \caption{Quantitative analysis of the different methods at two acceleration rates (4 and 8).}
        \begin{tabular}{l|l|l|ll}
        \hline
        \multirow{2}{*}{Methods}                    & \multicolumn{2}{l|}{PSNR} & \multicolumn{2}{l}{SSIM}                   \\ \cline{2-5} 
                                                    & 4$\times$   & 8$\times$   & \multicolumn{1}{l|}{4$\times$} & 8$\times$ \\ \hline
        U-Net-256                                   & 30.833      & 29.252      & \multicolumn{1}{l|}{0.89184}   & 0.85748   \\ \hline
        SSDU                                        & 35.908      & 32.469      & \multicolumn{1}{l|}{0.95130}   & 0.91531   \\ \hline
        Ours                                        & 38.575      & 33.255      & \multicolumn{1}{l|}{0.97177}   & 0.92709   \\ \hline
        Supervised                                  & 39.471      & 33.928      & \multicolumn{1}{l|}{0.97843}   & 0.93919   \\ \hline
    \end{tabular}
    \label{tab.1}
\end{table}

\subsubsection{Comparison Methods.}
Our proposed method is compared to the following reconstruction methods to evaluate the effectiveness: 1) U-Net-256: A U-Net model trained in a supervised manner, where the number of channels of the last encoding layer is 256. 2) SSDU: An ISTA-Net$^+$ model trained in a self-supervised manner as in \cite{self-supervised}. In this paper, a slight difference in the network loss calculation is made. To ensure stable network training, more k-space data points (including partial network input) are utilized to calculate the loss. 3) Supervised : An ISTA-Net$^+$ model trained in a supervised manner.

\begin{figure}[!htbp]
    \centering
    \includegraphics[width=\textwidth]{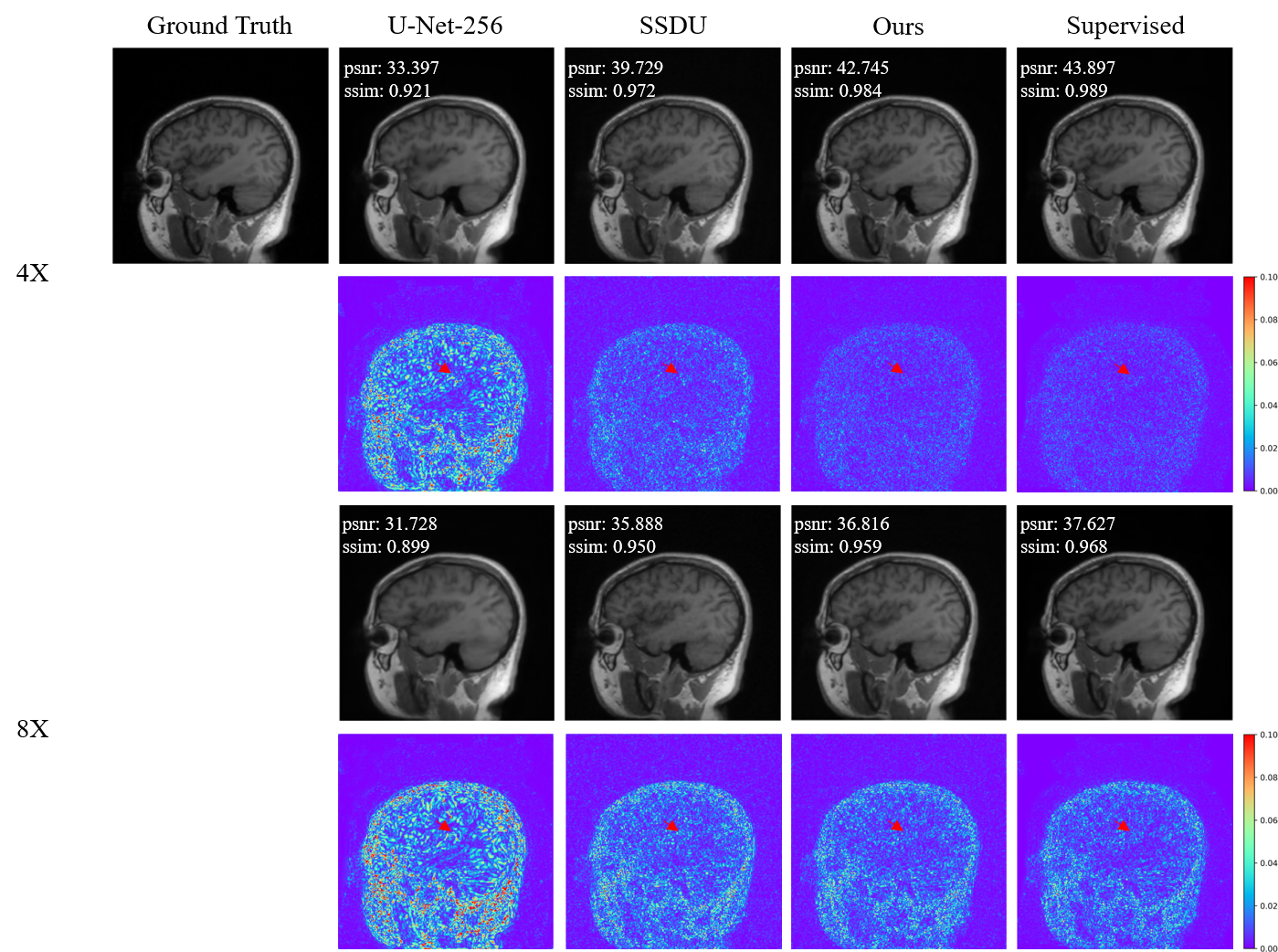}
    \caption{Example reconstruction results of the different methods at two acceleration rates (4 and 8) along with their corresponding error maps.}
    \label{fig.3}
\end{figure}

Table \ref{tab.1} lists the reconstruction results of the different methods. PSNR and SSIM represent the peak signal-to-noise ratio and the structural similarity index measurement values. Compared to U-Net-256 and SSDU, our method achieves significantly improved reconstruction performance at both acceleration rates. Moreover, our method generates very competitive results which are close to those generated by the corresponding supervised learning method that utilizes the exact same network architecture (``Supervised'' in Table \ref{tab.1}). Example reconstruction results are plotted in Fig. \ref{fig.3}. It can be observed that our method recovers more detailed structural information compared to U-Net-256 and SSDU, which suggests that our method can reconstruct MR images with better visual qualities.

\subsubsection{Ablation Analysis.}
To evaluate the effectiveness of the different components of the proposed method, ablation studies are performed and the results are reported in Table \ref{tab.2}. It can be summarized that employing the defined reconstruction loss (utilizing all the available undersampled data to calculate the reconstruction loss) yields significantly better performance compared to SSDU, where only part of the undersampled data is used to calculate the loss. Besides, as illustrated in Table \ref{tab.2}, with or without sharing the parameters of the parallel networks, the reconstruction performance is consistently improved when the difference loss is introduced, which suggests the effectiveness of the proposed difference loss for the high-quality image reconstruction task.
\begin{table}[!htbp]
    \centering
    \caption{Ablation study results at two acceleration rates (4 and 8). ``wo DiffLoss'' means without the difference loss. ``share'' represents sharing the parameters of the two parallel networks.}
        \begin{tabular}{l|l|l|ll}
            \hline
            \multirow{2}{*}{Methods}                & \multicolumn{2}{l|}{PSNR}             & \multicolumn{2}{l}{SSIM}                                  \\ \cline{2-5} 
                                                    & 4$\times$         & 8$\times$         & \multicolumn{1}{l|}{4$\times$}        & 8$\times$         \\ \hline
            SSDU                                    & 35.908            & 32.469            & \multicolumn{1}{l|}{0.95130}          & 0.91531           \\ \hline
            Self-supervised / wo DiffLoss           & 38.054            & 33.077            & \multicolumn{1}{l|}{0.97036}          & 0.92439           \\ \hline
            Self-supervised / share                 & 38.353            & 33.216            & \multicolumn{1}{l|}{0.97086}          & 0.92681           \\ \hline
            Self-supervised / wo share (Ours)       & \textbf{38.575}   & \textbf{33.255}   & \multicolumn{1}{l|}{\textbf{0.97177}} & \textbf{0.92709}  \\ \hline
        \end{tabular}
    \label{tab.2}
\end{table}
\section{Conclusion}
A novel self-supervised learning method for MRI reconstruction is proposed that can be employed by any existing deep learning-based reconstruction model. With our method, neural networks obtain the ability to infer unknown frequency information, and thus, high-quality MR images can be reconstructed without utilizing any fully sampled reference data. Extensive experimental results confirm that our method achieves competitive reconstruction performance when compared to the corresponding supervised learning method at high acceleration rates.

\subsubsection{Acknowledgments.}
This research was partly supported by the National Natural Science Foundation of China (61871371, 81830056), Key-Area Research and Development Program of GuangDong Province (2018B010109009), Scientific and Technical Innovation 2030-``New Generation Artificial Intelligence" Project (2020AAA0104100, 2020AAA0104105), Key Laboratory for Magnetic Resonance and Multimodality Imaging of Guangdong Province (2020B1212060051), the Basic Research Program of Shenzhen (JCYJ20180507182400762), Youth Innovation Promotion Association Program of Chinese Academy of Sciences (2019351).

\bibliographystyle{splncs04}
\bibliography{paper1698}

\end{document}